\pdfoutput=1

\documentclass[11pt]{article}

\usepackage{acl}

\usepackage{times}
\usepackage{latexsym}
\usepackage{url}

\usepackage{amsmath}
\usepackage{amssymb}
\usepackage{bm}
\usepackage{algorithm}
\usepackage[noend]{algpseudocode}
\usepackage{multirow}
\usepackage{booktabs} 
\usepackage{tabularx}
\usepackage{graphicx} 
\usepackage{tablefootnote}

\usepackage{listings,multicol}

\usepackage{xcolor}
\definecolor{codegreen}{rgb}{0,0.6,0}
\definecolor{codegray}{rgb}{0.5,0.5,0.5}
\definecolor{codepurple}{rgb}{0.58,0,0.82}
\definecolor{backcolour}{rgb}{0.95,0.95,0.92}

\usepackage[T1]{fontenc}

\usepackage[utf8]{inputenc}

\usepackage{microtype}

\title{ \textsc{SciLit}: A Platform for Joint Scientific Literature Discovery, Summarization and Citation Generation }

\author{Nianlong Gu \\
  Institute of Neuroinformatics,\\ 
  University of Zurich and\\
  ETH Zurich\\
  \texttt{nianlong@ini.ethz.ch} \\\And
  Richard H.R. Hahnloser\\
  Institute of Neuroinformatics,\\
  University of Zurich and\\
  ETH Zurich\\
  \texttt{rich@ini.ethz.ch}\\
  }

\begin{document}
\maketitle
\begin{abstract}

Scientific writing involves retrieving, summarizing, and citing relevant papers, which can be time-consuming processes in large and rapidly evolving fields. By making these processes inter-operable, natural language processing (NLP) provides opportunities for creating end-to-end assistive writing tools.  We propose \textsc{SciLit}, a pipeline that automatically recommends relevant papers, extracts highlights, and suggests a reference sentence as a citation of a paper, taking into consideration the user-provided context and keywords.
\textsc{SciLit} efficiently recommends papers from large databases of hundreds of millions of papers using a two-stage pre-fetching and re-ranking literature search system that flexibly deals with addition and removal of a paper database. 
We provide a convenient user interface that displays the recommended papers as extractive summaries and that offers abstractively-generated citing sentences which are aligned with the provided context and which mention the chosen keyword(s).
Our assistive tool for literature discovery and scientific writing is available at \url{https://scilit.vercel.app}

\end{abstract}

\section{Introduction}

When we compose sentences like ``Our experiments show that XXX performs significantly worse than YYY'' in a manuscript, we may want to find papers that report similar performance evaluations \cite{cohan-etal-2019-structural} and discuss these in our manuscript. 
This process is a non-trivial task requiring in-depth human involvement in finding, summarizing, and citing papers, which raises the question whether it is possible to partly automate this process to reduce users' cognitive load in searching, retrieving, reading, and rephrasing related findings.

\begin{figure}
\centering
  \includegraphics[width=\linewidth]{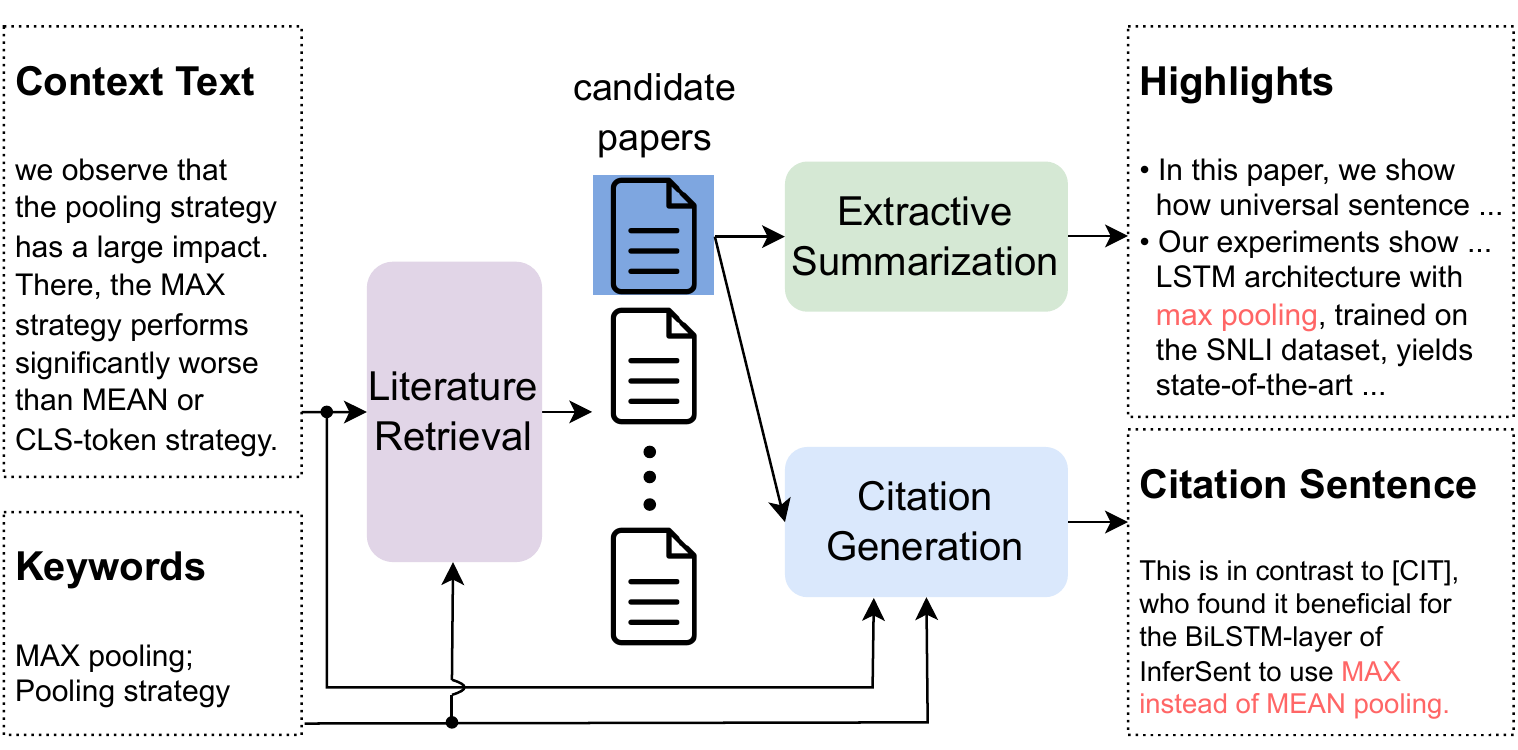}
  \caption{
  The main workflow of our platform. The example text for the context was selected from \citet{reimers2019sentence}. 
  }
  \label{fig:pipeline-architecture}
\end{figure}

Recent advances in natural language processing (NLP) help answer this question. First, releases of large scientific corpora such as S2ORC \cite{lo-wang-2020-s2orc} and  General Index \cite{else2021giant} provide opportunities for building large databases of scientific papers.
Second, such databases can be linked to systems for text retrieval \cite{guo2020deep}, citation recommendation \cite{farber2020citation, gu_local_2022,medic-snajder-2020-improved}, extractive summarization \cite{zhong_extractive_2020,9257174,gu-etal-2022-memsum}, and citation generation  \cite{xing_automatic_2020,ge_baco_2021,wang_disencite_2022}, all of which can be tailored to meet the requirements of an author's manuscript. 

To build a comprehensive system that helps authors with finding, reading, and summarizing of literature, the following challenges must be overcome: The system must index many papers (e.g., S2ORC has over 136 million papers \cite{lo-wang-2020-s2orc}) to achieve good coverage, it must respond quickly to queries, and it must be flexible to handle database additions and deletions. In addition, the overall architecture should be modular so that components can be easily upgraded when better algorithms become available.

To meet these challenges, we developed \textsc{SciLit}, a platform for literature discovery, summarization, and citation generation. We propose a hierarchical architecture for paper retrieval that efficiently retrieves papers from multiple large corpora. On each corpus (e.g., S2ORC and PMCOA \cite{pmcoa}), we build an efficient prefetching system based on a keyword inverted index and a document embedding index. The prefetched documents are then re-ordered (re-ranked) by a fine-tuned SciBERT \cite{beltagy-etal-2019-scibert}. Such an architecture allows us to dynamically add and remove databases and update one database and its index without significantly affecting the others. From a retrieved document (i.e., target paper), we extract highlights using a light-weight extractive summarization model proposed in \citet{gu-etal-2022-memsum}. Furthermore, using a fine-tuned T5 model \cite{2020t5}, we generate a citing sentence based on the abstract of the target paper, the context (the text surrounding the original citation sentence), and the keywords provided by a user. We also develop a microservice-based architecture that allows easy updating of algorithms.

In summary, our main contributions are: \begin{itemize}
    \item
    We demonstrate \textsc{SciLit}, a platform for searching, summarizing, and citing scientific papers.
    
    \item 
    We evaluate \textsc{SciLit} on scientific literature retrieval, paper summarization, and context-aware citation sentence generation, and showcase the generation of a related-work paragraph.
    
    \item A live demo website of our system is at \url{https://scilit.vercel.app}~and our implementation and data are at \url{https://github.com/nianlonggu/SciLit} and a video demonstrating the system can be viewed at \url{https://youtu.be/PKvNaY5Og1Y}
    
\end{itemize}

\section{SciLit}
Figure~\ref{fig:pipeline-architecture} shows the workflow of our system. A literature discovery module receives a context (a text) and keywords provided by a user and recommends a list of relevant papers that are semantically similar with the context and that match the keywords used as a Boolean filter \cite{gokce-etal-2020-embedding}. For each recommended paper, an extractive summarizer selects a short list of sentences from the full text as highlights. From the target paper selected by the user, a citation generation module takes the abstract together with the context and keywords as inputs and generates a citation sentence that references the target paper and that fits the context and keywords.

We define the context as the text before a citation sentence because we focus on the workflow of first finding papers (i.e., the missing citation as in as in \citet{gu_local_2022,medic-snajder-2020-improved}) and then writing citation sentences. The user-provided keywords are optional. When no keywords are explicitly given during training and evaluation of our system, we extract keywords from the context, the cited paper, and the citation sentence as substitutes.

\begin{figure}
\centering
  \includegraphics[width=\linewidth]{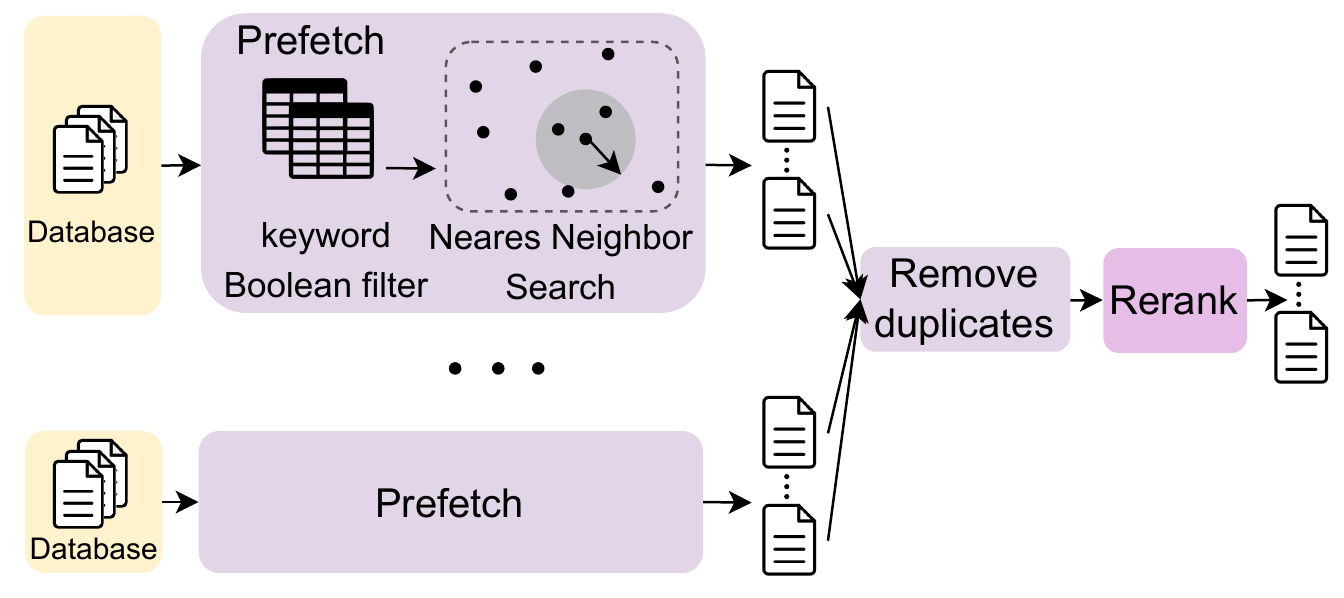}
  \caption{Schematic of literature discovery (paper retrieval). From each database, candidate documents are prefetched by a cascade of keyword Boolean filter and embedding-based nearest neighbor search. Then, candidate documents are reranked by a fine-tuned SciBERT.}
  \label{fig:literature-retrieval}
\end{figure}

\subsection{Literature Discovery}

\begin{table*}
\centering
\resizebox{\linewidth}{!}{ 
\begin{tabular}{lccccccccc}
\toprule 
\multirow{2}*{\shortstack{$ $\\\textbf{Corpus}}}  &
\multicolumn{3}{c}{\textbf{Databases}}& \multicolumn{4}{c}{\textbf{Inverted Index}} & \multicolumn{2}{c}{\textbf{Embedding Index}} \\
\cmidrule(lr){2-4}
\cmidrule(lr){5-8}
\cmidrule(lr){9-10}

& \shortstack{\# of\\papers} &\shortstack{\# papers with\\ fullbody} & \shortstack{until\\date} &  \shortstack{keywords\\ length} & \shortstack{\# of\\keywords} & \shortstack{data\\format} & \shortstack{storage\\size} & \shortstack{embedding\\dimension} & \shortstack{storage\\size}  \\
\midrule

S2ORC & 136.60 M & 12.44 M  & 2020-04-14  & \multirow{3}*{ \shortstack{unigram, bigram} }  & 1.20 B & \multirow{3}*{ sqlitedict 
} & 769 GB  & \multirow{3}*{ 256 } & 169 GB \\
PMCOA & 2.89 M & 2.73 M  & 2022-06-17 &  & 0.30 B &  & 145 GB &  & 2.9 GB \\
arXiv & 1.69 M & 1.69 M  & 2022-07-28 &  & 0.15 B &  & 77 GB &  & 1.7 GB \\

\bottomrule
\end{tabular}
}
\caption{ \label{tab:literature_discovery_statistics} 
Statistics of our literature discovery system. We indexed S2ORC  \cite{lo-wang-2020-s2orc}, PMCOA \cite{pmcoa}, and arXiv \cite{arxivdataset}, which contain large numbers of recent scientific papers in diverse fields.
}
\end{table*}

The literature discovery module takes as inputs the context and keywords and recommends papers that are worth citing in the context. To strike a balance between query accuracy and speed on large scientific corpora, our document discovery module uses a two-stage prefetching-ranking strategy \cite{gu_local_2022} (Figure \ref{fig:literature-retrieval}): For each scientific corpus, we build a database and create an efficient prefetching model that we use to pre-filter a number $N_p$ (see the discussion of $N_p$ in Table \ref{tab:retrieval_performance} and Section \ref{sec:performance}) of candidate documents from each corpus based on the provided keywords and context. After removing duplicates, we re-rank all prefetched documents using a trained reranker.

\noindent\textbf{Databases.} 
We dump each corpus into a separate SQLite \cite{sqlite2022hipp} database to allow flexibility in deploying and maintaining of independent prefetching servers. We further process documents from different corpora into a unified JSON schema compatible with a single codebase for indexing, querying, summarizing, and displaying documents from different corpora. The JSON schema includes keys ``Title'', ``Author'', etc., for parsed metadata, and ``Content.Abstract\_Parsed'', ``Content.Fullbody\_Parsed'' for parsed full text,
The details are given in Appendix \ref{sec:appendix-json-schema}.

\noindent\textbf{Prefetching.}
The prefetching model of a given SQLite database consists of an inverted index and an embedding index. The inverted index stores the paper IDs (unique identifiers for retrieving a paper's content) of all publications that contain a given keyword, e.g., a unigram such as ``computer'' or a bigram such as ``machine learning''. The embedding index is a table of embeddings of all papers in the database. Embeddings are 256-dimensional vectors computed by Sent2Vec \cite{pagliardini_unsupervised_2018} (we simply average the embeddings of all words in a document). We trained Sent2Vec on sentences from the full-text papers contained in S2ORC.

Using the keywords and a specific syntax, we first perform Boolean filtering \cite{gokce-etal-2020-embedding} of the inverted index. For example, given ``POS tag;2010..2022'', we filter papers published between 2010 and 2022 that mention ``POS tag''. 
The filtered papers are then ranked based on the cosine similarity between the papers' Sent2Vec embeddings and the context embedding.
Such a hybrid of lexical filtering and semantic ranking allows users to find papers that are semantically similar to the context and that flexibly meet a constrained search scope. 

Statistics for the database and indexing system are reported in Table \ref{tab:literature_discovery_statistics}. Details of the indexing implementation are shown in Appendix \ref{sec:appendix-prefetching-index}.

\noindent\textbf{Duplicate Removal.}
Since corpora can overlap, the prefetched candidates from multiple corpora can contain duplicate items. To remove duplicated candidates, we check the title and authors and keep only one record per paper for reranking.

\noindent\textbf{Reranking.} 
We use SciBERT \cite{beltagy-etal-2019-scibert} to rerank prefetched candidate papers, aiming at highly ranking candidates that can be cited given the context and keywords. We follow \citet{gu_local_2022} to compute an affinity score as follows: we pass an input text ``[CLS]\textit{query}[PAD]\textit{paper}[PAD]'' to SciBERT, where the \textit{query} $q$ is a concatenation of the context and the keywords, and \textit{paper} $d$ is a concatenation of the title and the abstract of the candidate paper. 
The encoded output of the ``[CLS]'' token is passed to a linear layer, which outputs a scalar $s(q,d)$ that we interpret as the affinity score between the query $q$ and the paper $d$.
To train the reranker, we use the cross-entropy loss:
\begin{equation}
    \begin{aligned}
    L = - \log \frac{ \exp{s(q, d^+)} }{ \exp{s(q, d^+)} + \sum_{i=1}^{N}{\exp{s(q, d^-_i)}} },
    \end{aligned}
\end{equation}
where $d^+$ is the paper actually cited in the query, and $d^-_i$ is one of $N (N=10)$ uncited papers that are randomly sampled from prefetched candidates at each training iteration.


\subsection{ Extractive Summarization}

The extractive summarization module extracts a short list of sentences from the full text of a paper to highlight the main points to a reader. We choose the summary to be extractive rather than abstractive to prevent readers from being misled by potential hallucinations introduced by abstractive summarization models
\cite{nan-etal-2021-entity,xu-etal-2020-fact,wang_asking_2020}. 
The extractive summarization model must efficiently select sentences from a given document so that users do not experience obvious delays.

In this paper, we employ MemSum, an RNN-based extractive summarizer that models the extraction process as a Markov decision process in a reinforcement learning framework. 
MemSum has been trained on the PubMed dataset \citet{gu-etal-2022-memsum} and it can summarize long papers without exhausting GPU memory due to its lightweight model structure. Also, MemSum is computationally efficient, taking only 0.1 sec on average to summarize a paper. These features make it a suitable model for our extractive summarization module. 

\subsection{ Citation Generation Module }
The citation generation module acts as an abstract summarizer that takes as input the context, the keywords, and the target paper to be cited; it then generates a sentence that cites the target paper and narrates it in context. 

By providing keywords as inputs to a sequence-to-sequence model, our input differs from previous works on automatic citation generation \cite{ge_baco_2021,xing_automatic_nodate}, which use only the context as inputs but no keywords. We consider keywords to be an important source of input because we believe that authors usually have a clear intention when citing a paper, and a keyword can sometimes more easily convey this intention than a long text. In the case shown in Figure \ref{fig:pipeline-architecture}, for example, after writing the context ``MAX pooling performs worse than MEAN pooling'', the author naturally intends to discuss papers about ``MAX pooling''. Therefore, the keyword ``MAX pooling'' should be used as a thematic cue for citation sentence generation. Moreover, making the citation generation model conditional on keywords also allows users to fine-tune the generated citation text by simply adjusting the keywords, thus making the system interactive and conveniently tunable.

To make the generation conditional on context, keywords, and cited papers, we fine-tuned a T5 \cite{2020t5} so that its input is a concatenation of three attributes: keywords, context, and the abstract of a cited paper, each preceded by a special field name to make attributes distinguishable to the model:
\noindent\texttt{keywords: XXX. context: XXX. target abstract: XXX.}
\noindent The corresponding decoding output is the actual citation sentence that cites the target paper. 


\subsection{Microservice-based Architecture}

\begin{figure}
\centering
  \includegraphics[width=\linewidth]{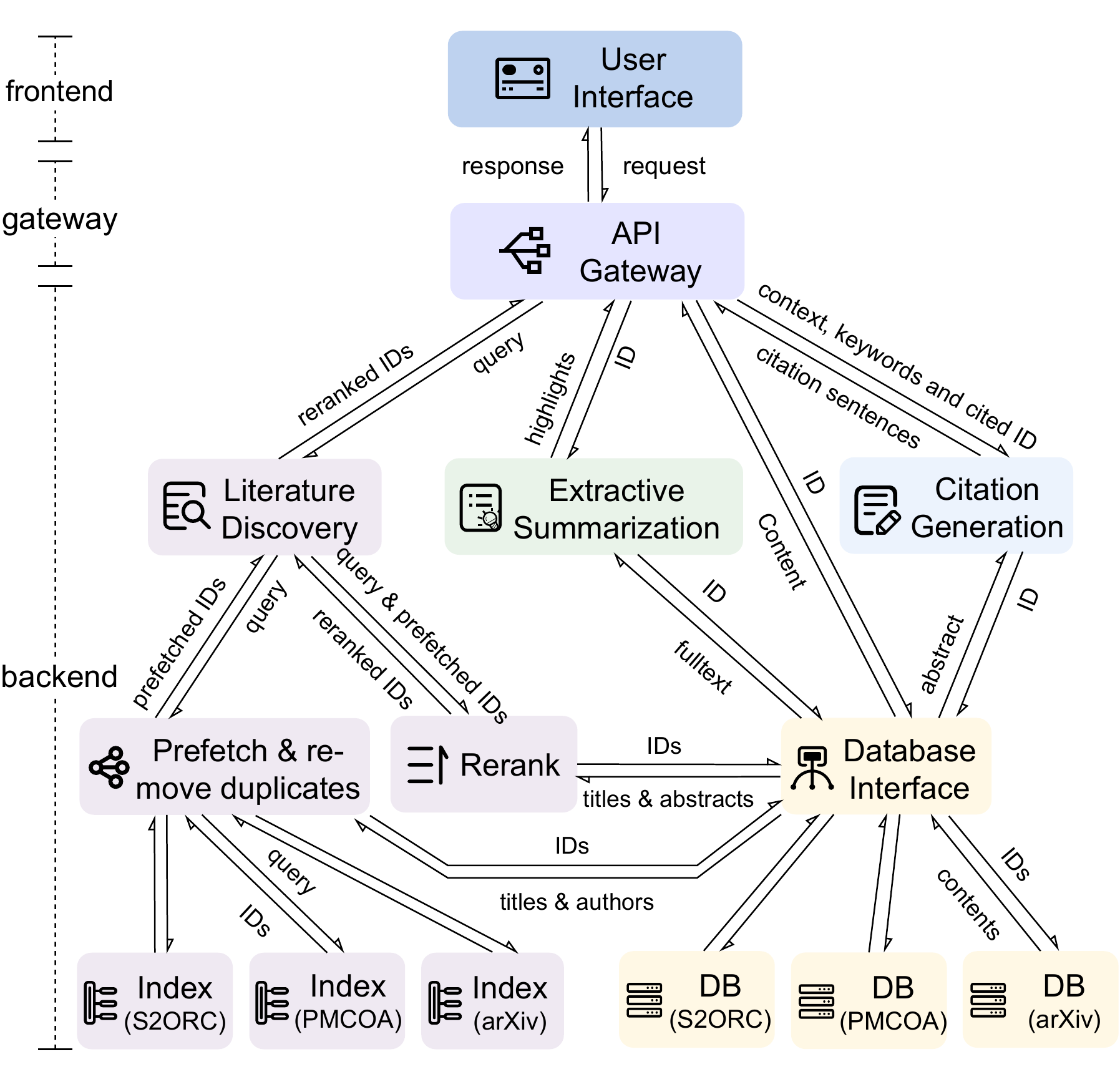}
  \caption{The architecture of our platform. The direction of an arrow represents the direction of data flow.
  }
  \label{fig:backend-overall-architecture}
\end{figure}

\begin{figure*}[htb]
\centering
  \includegraphics[width=\linewidth]{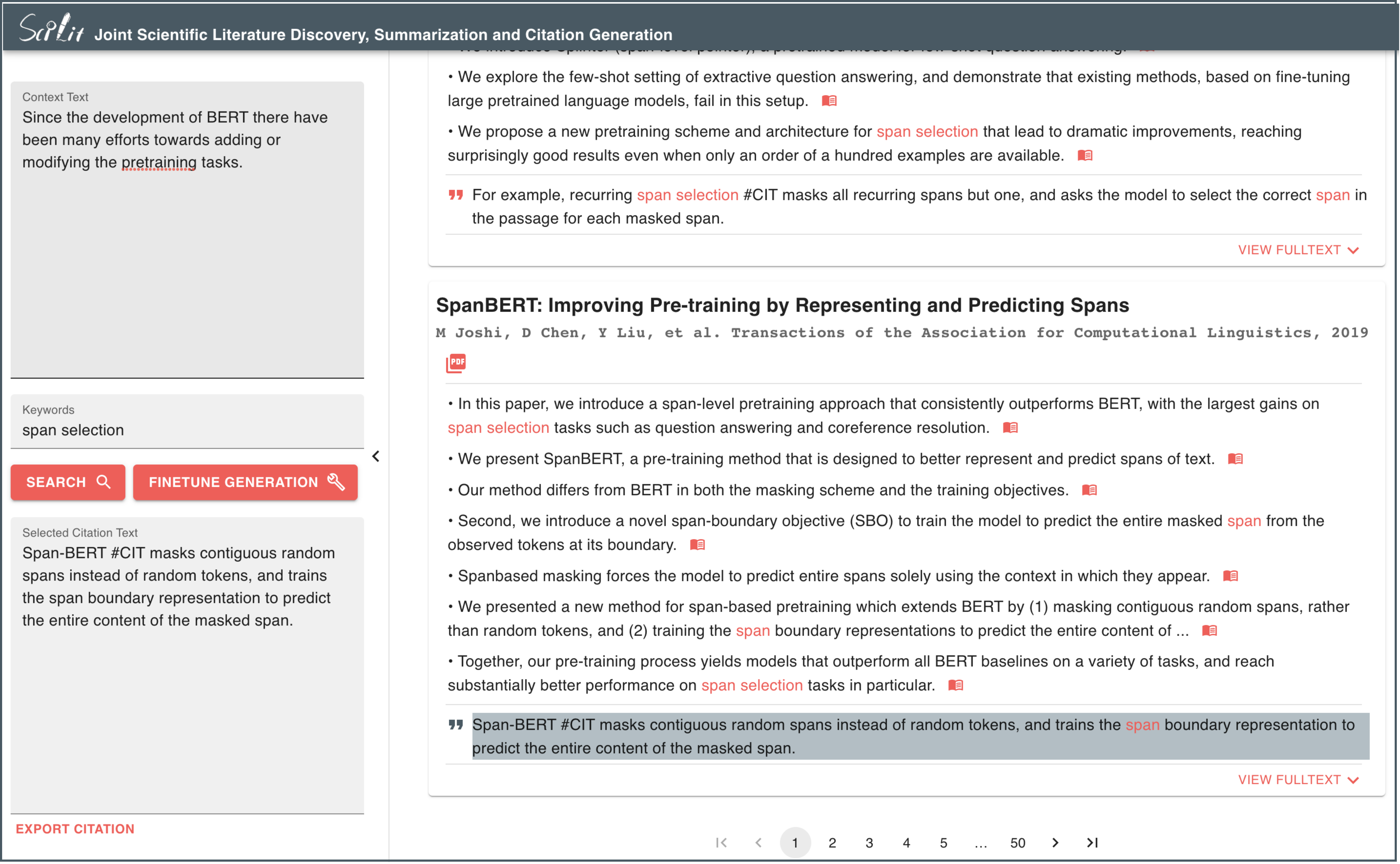}
  \caption{Overview of the user interface. The context text comes from the related work section in \citet{glass-etal-2020-span}.
  }
  \label{fig:user-interface-overview}
\end{figure*}

We built our platform as a network of microservices (Figure \ref{fig:backend-overall-architecture}). An API gateway routes requests from the frontend to the target microservice on the backend. The microservices run separate modules on their respective Flask servers \cite{aggarwal2014flask} and communicate with each other by sending HTTP requests and waiting for responses. When a query request arrives, the API gateway forwards the query to the literature discovery service, which calls the prefetching and reranking services to get the reranked IDs. The API gateway then sends the paper IDs to the extractive summarization service to receive the highlights of each recommended paper. The gateway also sends the context, keywords, and recommended paper IDs to the citation generation service to suggest citation sentences. The database interface service manages the databases of multiple scientific corpora and provides a unified interface to access the paper content given its ID. Each microservice runs in an independent environment, which makes it easy to upgrade backend systems online, such as adding or removing a database or updating an algorithm.

\section{Evaluation}
In this section, we first show how \textsc{SciLit} works and then we evaluate its performance.
\subsection{Demonstration}
Our user interface runs on a web page (Figure \ref{fig:user-interface-overview}) created with ReactJS\footnote{\url{https://reactjs.org/}}. The left sidebar is an input panel where users can enter context and keywords and trigger a query by clicking the \textit{search} button. Retrieved papers are displayed in the search-results panel on the right. Users can scroll up and down or paginate to browse through the recommended papers. Each paper is accompanied by highlights and a suggested citation sentence generated by our extractive summarization and citation generation services, respectively. Users can cite a paper by clicking on the \textit{cite} button and the suggested citation sentence will jump to the editing area on the left where users can tweak the sentence by changing keywords and clicking on the \textit{fine-tune generation} button, or they can edit the sentences manually. Exporting citation information is also supported.
\subsection{Performance}
\label{sec:performance}
\noindent\textbf{Evaluation Dataset.} 
We evaluated \textsc{SciLit} on a test set containing $1530$ samples, which are mainly papers published in 2022 in the fields of computer and biomedical sciences. 
Each sample contains the following information: 1) context, up to $6$ sentences preceding the citation sentence and within the same section; 2) keywords, up to $2$ uni- or bi-grams that occur in all of the context, the citation sentence, and the cited paper; 3) ID of the cited paper; 4) the citation sentence following the context, which is the ground truth for evaluating generated citations. For quality control, we only include citation sentences in the test set that cite one paper.

\begin{table}
\centering
\resizebox{\linewidth}{!}{ 
\begin{tabular}{lccccccc}
\toprule 
\multirow{2}*{$N_p$}& \multirow{2}*{\shortstack{time\\($s$/query)}} & \multicolumn{6}{c}{Recall@$K$ (R@$K$) } \\ \\[-1.2em]
\cmidrule(lr){3-8} \\[-1.5em]
& & R@1 & R@5 & R@10 & R@20 & R@50 & R@100 \\ 
\midrule \\[-1em]
50 & \textbf{2.02} &\textbf{0.107} & 0.208 & 0.263 & 0.305 & 0.327 & 0.331 \\
100 & 2.55 &0.096 & 0.215 & \textbf{0.278} & 0.328 & 0.384 & 0.401 \\
200 & 3.26 &0.095 & \textbf{0.220} & 0.275 & \textbf{0.339} & 0.420 & 0.452 \\
300 & 3.93 &0.095 & 0.204 & 0.273 & 0.330 & \textbf{0.422} & \textbf{0.482} \\
\bottomrule
\end{tabular}
}
\caption{ \label{tab:retrieval_performance} Paper retrieval performance measured as the recall of the top $K$ recommendations. $N_p$ denotes the number of prefetched candidates per corpus.
}
\end{table}
\noindent\textbf{Paper Retrieval.}
For each sample in the evaluation dataset, we use context and keywords as queries and invoke the literature search service to first prefetch $N_p$ candidates from each of the three corpora (S2ORC, PMCOA, and arXiv). We remove duplicates and then we rank the prefetched candidates. The top $K$ recommendations serve to evaluate the retrieval performance (Table \ref{tab:retrieval_performance}).  We observed that for large $K (K=50, 100)$, the recall increases as $N_p$ increases, whereas for small $K (K=5,10,20)$, the recall first increases and then starts to decrease, indicating that the reranking performance is impacted by increasing number of prefetched candidates. We choose $N_p=100$ as the default value, which gives rise to fast reranking and achieved the best performance at R@10.

\begin{table}
\centering
\resizebox{\linewidth}{!}{ 
\begin{tabular}{lccc}
\toprule 
Model & Rouge-1 & Rouge-2 & Rouge-L\\ 
\midrule \\[-1em]
BertSum \cite{https://doi.org/10.48550/arxiv.1903.10318} & 42.53 & 16.89 & 39.18 \\
MemSum \cite{gu-etal-2022-memsum} & \textbf{46.40}* & \textbf{19.61}* & \textbf{42.66}* \\ 
\bottomrule
\end{tabular}
}
\caption{ \label{tab:extractive_summarization_performance} The extractive summarization performance. 
“*" indicates statistical significance in comparison to baselines with a 95\% bootstrap
confidence interval.
}
\end{table}

\noindent\textbf{Extractive Summarization.} 
To evaluate the summaries, following \citet{zhong_extractive_2020,xiao-carenini-2019-extractive}, we computed the ROUGE F1 scores between the summary sentences extracted from the full body and the corresponding abstract.
MemSum significantly outperformed BertSum \cite{https://doi.org/10.48550/arxiv.1903.10318}, a Bert-based summarizer that requires truncation of long documents, indicating the effectiveness of MemSum in extractively summarizing scientific documents.

\begin{table}
\centering
\resizebox{\linewidth}{!}{ 
\begin{tabular}{lccc}
\toprule 
generation pipeline $\ \ \ $ & Rouge-1 & Rouge-2 & Rouge-L\\ 
\midrule \\[-1em]
generation-only $\ \ \ $& 32.96 & 9.19 & 24.52 \\
Best of top $1$ paper $\ \ \ $& 28.62 & 6.00 & 21.05 \\
Best of top $5$ papers $\ \ \ $& 34.92 & 9.59 & 26.23 \\
Best of top $10$ papers $\ \ \ $& \textbf{36.83}* & \textbf{10.98}* & \textbf{28.10}* \\
\bottomrule
\end{tabular}
}
\caption{ \label{tab:citation_generation_performance} The citation generation performance.
}
\end{table}

\noindent\textbf{Citation Generation.} 
We assessed our joint retrieval and citation generation system by letting it recommend papers based on context and keywords first, then generate $K$ citation sentences corresponding to each of the top $K$ suggested papers. Next, we calculate the ROUGE F1 score by comparing each generated citation sentence with the ground-truth citation sentence (the sentence that actually follows the context text in the test example), keeping track of the highest ROUGE F1 score in this set. This method is named the "Best-of-top-$K$" procedure for ease of understanding.

We hypothesized that generating multiple citation sentences, one for each top-$K$ paper, would increase the chances of crafting a suitable citation sentence. To verify this, we compared the effectiveness of the "Best-of-top-$K$" approach with a "generation-only" system. The latter generates a single citation sentence based on the actual cited paper provided as ground truth.

We observed that for $K=5$ and $10$, the ``Best-of-top-$K$'' pipeline achieved significantly higher ROUGE scores than the "generation only" pipeline (Table \ref{tab:citation_generation_performance}), indicating that the paper retrieval module contributes positively to the citation generation process and increases the chance of suggesting appropriate citation sentences. We believe that this result further supports our idea of developing an integrated system for joint retrieval and generation.

\subsection{Ablation Study}

\begin{table}
\centering
\resizebox{\linewidth}{!}{ 
\begingroup

\setlength{\tabcolsep}{4pt}
\renewcommand{\arraystretch}{1}
\begin{tabular}{lcccccc}
\toprule 
\multirow{2}*{ \shortstack{retrieval$\ \ \ $\\($N_p=100$)}} & \multirow{2}*{R@1} & \multirow{2}*{R@5} &\multirow{2}*{R@10} & \multirow{2}*{R@20} & \multirow{2}*{R@50} & \multirow{2}*{R@100} \\ \\
\midrule
w keywords &0.096 & 0.215 & 0.278 & 0.328 & 0.384 & 0.401 \\
w/o keywords & 0.013 & 0.050 & 0.085 & 0.125 & 0.199 & 0.250 \\
\bottomrule
\toprule
\multicolumn{2}{l}{citation generation} & Rouge-1 & \multicolumn{2}{r}{Rouge-2} & \multicolumn{2}{c}{ Rouge-L} \\ 
\midrule
 \multicolumn{2}{l}{w keywords} &32.96 & \multicolumn{2}{r}{9.19$\ \ \ $} & \multicolumn{2}{c}{24.52}  \\
\multicolumn{2}{l}{w/o keywords} & 26.57 & \multicolumn{2}{r}{5.56$\ \ \ $} & \multicolumn{2}{c}{20.39}  \\
\bottomrule
\end{tabular}

\endgroup

}
\caption{ \label{tab:ablation-study} Ablation study on retrieval and citation generation performance.
}
\end{table}

To analyze the impact of keywords, we evaluated retrieval and generation systems without keywords. For document retrieval, we first prefetched $N_p=100$ candidates from each corpus and then ranked them based on context only. For citation generation, we trained a T5 model to learn to generate citation sentences with only the context and the title and abstract of the cited paper and evaluated it on the evaluation set. 
We observe a significant degradation in the performance of literature retrieval and citation generation (Table \ref{tab:ablation-study}), which demonstrates the utility of keywords for recommending relevant papers and generating accurate citations on our platform.


\section{Related Work}

Recently, AI-driven platforms focused on literature recommendation and scientific paper summarization have been proposed.
(keywords: \texttt{platform}, paper: \#2) \emph{One such platform is AI Research Navigator \cite{fadaee-etal-2020-new}, which combines classical keyword search with neural retrieval to discover and organize relevant literature.}
(keywords: \texttt{scientific;} \texttt{summarization;} \texttt{platform}, paper \#3) \emph{Another platform is Anne O'Tate, which supports user-driven summarization, drill-down and mining of search results from PubMed, the leading search engine for biomedical literature \cite{smalheiser2021anne}.} (keywords: \texttt{related work generation}, paper \#9) \emph{
\citet{https://doi.org/10.1002/cpe.4261} automatically generates related work by comparing the main text of the paper being written with the citations of other papers that cite the same references.}


In the previous paragraph, the italicized citation sentences are generated by \textsc{SciLit}. For generating a sentence, it used all preceding sentences in the paragraph as contexts and the keywords in parentheses. We browsed the 100 recommended papers by turning pages and reading the corresponding citation sentences (as shown in Figure \ref{fig:user-interface-overview}).  The numbers in parentheses indicate the ranks of the recommended papers.

\section{Conclusion and Future Work}

This paper demonstrates \textsc{SciLit}, a platform for joint scientific literature retrieval, paper summarization, and citation generation. \textsc{SciLit} can efficiently recommend papers from hundreds of millions of papers and proactively provide highlights and suggested citations to assist authors in reading and discussing the scientific literature.
In addition, our prefetching, reranking, and citation generation system can be conditioned on user-provided keywords, which provides flexibility and adjusts the platform's response to user intention. In the future, we will further improve the performance of each module, especially the citation generation part, and collect feedback from users to improve the overall workflow and the frontend user experience.

\section*{Acknowledgements}
This project was supported by the Open Research Data Program of the ETH Board (ORD2000103). We thank the anonymous reviewers for their useful comments. 

\bibliography{anthology,custom}
\bibliographystyle{acl_natbib}

\appendix

\section{Hardware Information}
\label{sec:appendix-hardware}
We run the backend of \textsc{SciLit} on a server with dual 64-Core AMD EPYC 7742 2.25GHz Processors, 2TB DDR4 3200MHz ECC Server Memory, and 4$\times$7.68TB NVME GEN4 PM9A3 for storage. The server is also equipped with two nVidia RTX A6000 48GB GPU. The frontend is hosted on Vercel\footnote{\url{https://vercel.com/}}.

\section{JSON Schema for Database}
\label{sec:appendix-json-schema}
The details of our unified JSON schema is shown in Listing \ref{code1}.  As metadata, we define the keys ``Author'', ``Title'', ``Abstract'', ``Venue'', ``DOI'', ``URL'', and as parsed full text we define the keys ``PublicationDate'',  "Content.Abstract\_Parsed", and ``Content.Fullbody\_Parsed''. The parsed abstract or full body contains a list of parsed sections. Each section contains a list of parsed paragraphs, each including a list of parsed sentences. If a sentence cites a paper, we create a ``cite span'' that records the citation marker such as ``[1]'', the position of the citation marker in the sentence, and the cited paper's index in the ``Reference'' list. 

We implemented a S2ORC parser to convert documents in the S2ORC corpus to our JSON format. For PDFs in the arXiv corpus, we first used the s2orc-doc2json \cite{lo-wang-2020-s2orc} to convert them into S2ORC format and then we applied our S2ORC parser. For XML files in the PMCOA corpus, we implemented an XML parser based on \citet{Achakulvisut2020} to convert XML to S2ORC format and then we applied again the S2ORC parser to convert it into our JSON format.

\lstdefinestyle{mystyle}{
  backgroundcolor=\color{backcolour}, commentstyle=\color{codegreen},
  keywordstyle=\color{magenta},
  numberstyle=\tiny\color{codegray},
  stringstyle=\color{codepurple},
  basicstyle=\ttfamily\footnotesize,
  breakatwhitespace=false,         
  breaklines=true,                 
  captionpos=b,                    
  keepspaces=true,                 
  numbers=left,                    
  numbersep=5pt,                  
  showspaces=false,                
  showstringspaces=false,
  showtabs=false,                  
  tabsize=2
}

\lstset{style=mystyle}
\begin{lstlisting}[language=Python, float=*, label=code1,caption=An example of the JSON schema that we used for parsing and storing scientific papers.]
{'Author': [{'GivenName': 'Daisuke', 'FamilyName': 'Ida'}, ...],
 'Title': 'Topology Change of Black Holes',
 'Abstract': 'The topological structure of the event horizon has been investigated ...',
 'Venue': '',
 'DOI': '',
 'URL': '',
 'PublicationDate': {'Year': '2007', 'Month': '3'},
  'Content': {
     'Abstract': '',
     'Abstract_Parsed': [{
         'section_id': '0',
         'section_title': 'Abstract',
         'section_text': [{
             'paragraph_id': '0',
             'paragraph_text': [{
                 'sentence_id': '0',
                 'sentence_text': 'The topological structure of the event horizon has been investigated in terms of the Morse theory.',
                 'cite_spans': []},
                 # ...
             ]},
             # ...
         ]
     }],
     'Fullbody': '',
     'Fullbody_Parsed': [{
         'section_id': '0',
         'section_title': 'Introduction',
         'section_text': [{
             'paragraph_id': '0',
             'paragraph_text': [
                 # ...,
                 {
                 'sentence_id': '2',
                 'sentence_text': '[1, 2] This follows from the fact that the total curvature, which is the integral of the intrinsic scalar curvature over the horizon, is positive under the dominant energy condition and from the Gauss-Bonnet theorem.',
                 'cite_spans': [{'start': '4', 'end': '6', 'text': '2]','ref_id': '0'}]
                 },
                 # ...
             ]
         }]
     }]},
 'Reference': [{
     'Title': 'The large scale structure of space-times',
     'Author': [{'GivenName': 'S', 'FamilyName': 'Hawking'},
                {'GivenName': 'G', 'FamilyName': 'Ellis'}],
     'Venue': '',
     'PublicationDate': {'Year': '1973'},
     'ReferenceText': '2. Hawking, S, and G Ellis. "The large scale structure of space-times." (1973).'},
     # ...
     ]
}
\end{lstlisting}

\lstset{style=mystyle}
\begin{lstlisting}[language=Python, float=*, label=code2,caption=The dictionary representation of the tree structure shown in Figure \ref{fig:appendix-syntax-parsing}.]
{
    'operation': 'AND',
    'elements': [
        {'operation': 'AND',
         'elements': [{'operation': None, 'elements': ['nlp']}]},
        {'operation': 'OR',
         'elements': [
            {'operation': 'AND',
             'elements': [{'operation': None, 
                           'elements': ['machine translation']}]},
            {'operation': 'AND',
             'elements': [{'operation': None, 'elements': ['nmt']}]}]},
        {'operation': 'OR',
         'elements': [
            {'operation': None, 'elements': ['publicationdate.year:2020']},
            {'operation': None, 'elements': ['publicationdate.year:2021']},
            {'operation': None, 'elements': ['publicationdate.year:2022']}
         ]
        }]
}
\end{lstlisting}

\section{Prefetching Indexing Implementation}
\label{sec:appendix-prefetching-index}

\subsection{Inverted Index}

The inverted index is a mapping table from keywords (unigrams and bigrams) to paper IDs. We extracted keywords from the full text of each document and kept a bigram only if neither word is a stopword. We use sqlitedict\footnote{\url{https://github.com/RaRe-Technologies/sqlitedict}} to store the inverted index for each corpus, which is an on-disk hashmap based on an SQLite database that allows us to efficiently obtain the paper ID for a given keyword without loading the entire inverted index into RAM.

\noindent\textbf{Syntax Parsing.} 
\begin{figure}
\centering
  \includegraphics[width=0.8\linewidth]{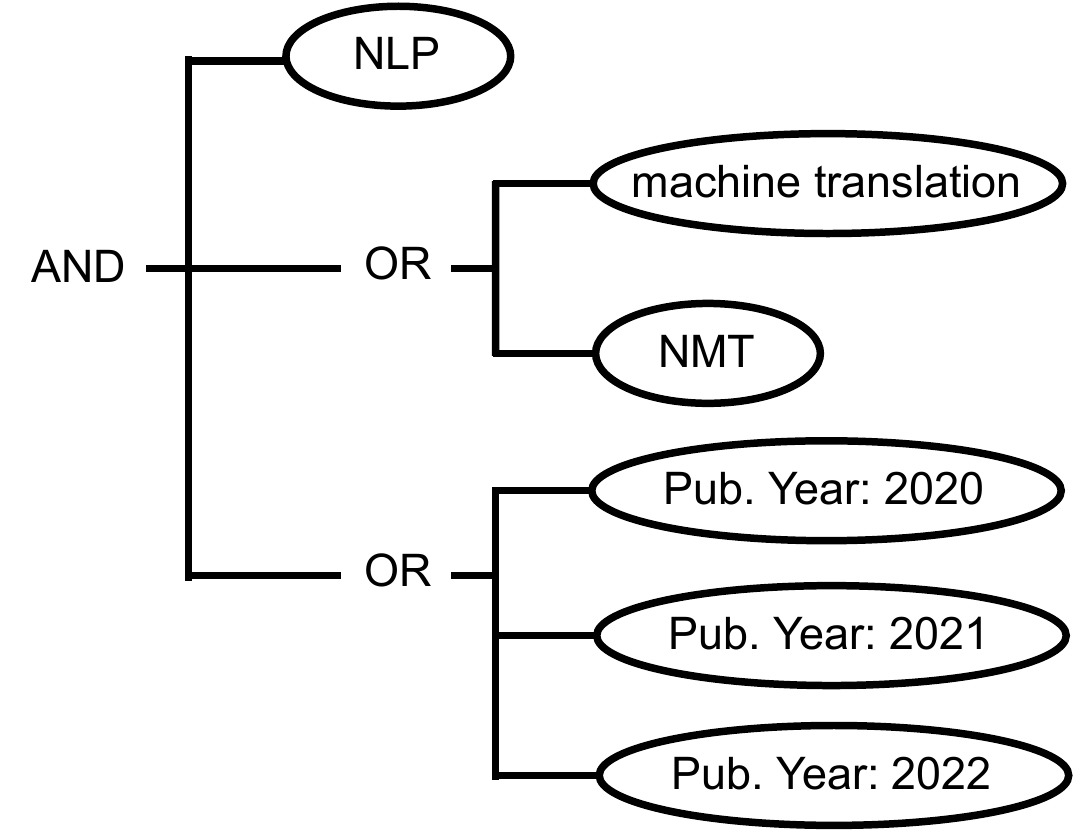}
  \caption{The parsed tree structure of the given keywords string: \texttt{NLP; machine learning|NMT; 2020..2022''}.
  }
  \label{fig:appendix-syntax-parsing}
\end{figure}
Our platform allows users to filter documents using syntax-rich keyword strings. For example, to filter papers that contain the keywords 'NLP' and either 'machine translation' or 'NMT' and that have been published between 2020 and 2022, one can compose a keyword string \texttt{NLP; machine learning|NMT; 2020..2022''}. We transform this keyword string into a tree of logical operations (Figure \ref{fig:appendix-syntax-parsing}), wherein each node defines the logical operation applied to the sub-nodes, and each leaf node contains a keyword. We implemented the tree using a Python dictionary (Listing \ref{code2}). We recursively traversed all nodes in the tree in a depth-first search, obtaining the paper IDs with the keyword in each leaf node, and applying the logical operations indicated in each node to obtain the final paper IDs at the root node.

\subsection{Embedding Index}
\noindent\textbf{Structure of the Embedding Index.} The embedding index consists of three main components:

The first component is a matrix $\mathcal{M}\in \mathcal{R}^{N\times D}$, where $N$ is the number of documents and $D$ is the dimensionality of document embeddings. Each document embedding is L2-normalized so that given an L2-normalized query embedding $e_q\in \mathcal{R}^{D\times 1}$, the matrix multiplication $\mathcal{M}e_q \in \mathcal{R}^{N\times 1}$ represents the cosine similarity between the query embedding and all document embeddings. We use $\mathcal{M}e_q$ to rank documents and to obtain the indices of the top-$K$ nearest neighbor (kNN) documents.

The second component is a mapping table from the index of a paper embedding in the matrix $\mathcal{M}$ to the corresponding paper ID in our databases. Using this mapping table, we retrieve a paper's content.

The last component is a reversed mapping table from the paper ID to the corresponding index in the embedding matrix. In our prefetching system, we first use the inverted index to pre-filter a subset of paper IDs based on given keywords. Then we use this reversed mapping table to obtain the corresponding paper embeddings and to perform KNN search among them.

\noindent\textbf{Multi-Processing Speedup for Brute-Force Nearest Neighbor Search.} 
For a large corpus like S2ORC, the embedding matrix contains up to 136.6 million vectors and so to perform matrix multiplication in a single thread is very time consuming. To take full advantage of the multiple CPU cores in our server, we sliced the embedding matrix into 137 shards, each containing about 1 million embeddings. On each document search, we first ran parallel brute-force nearest neighbor searches on each shard to obtain $N_p$ candidates each, and then we ranked the $137\times N_p$ candidates again to obtain the final $N_p$ candidates. Given that our server has 128 cores, we achieved a nearly linear speedup using such multithread KNN, which is mathematically equivalent to performing a single KNN over the entire embedding matrix.

\section{Joint Retrieval and Citation Generation Examples}
\begin{figure*}
\centering
  \includegraphics[width=\linewidth]{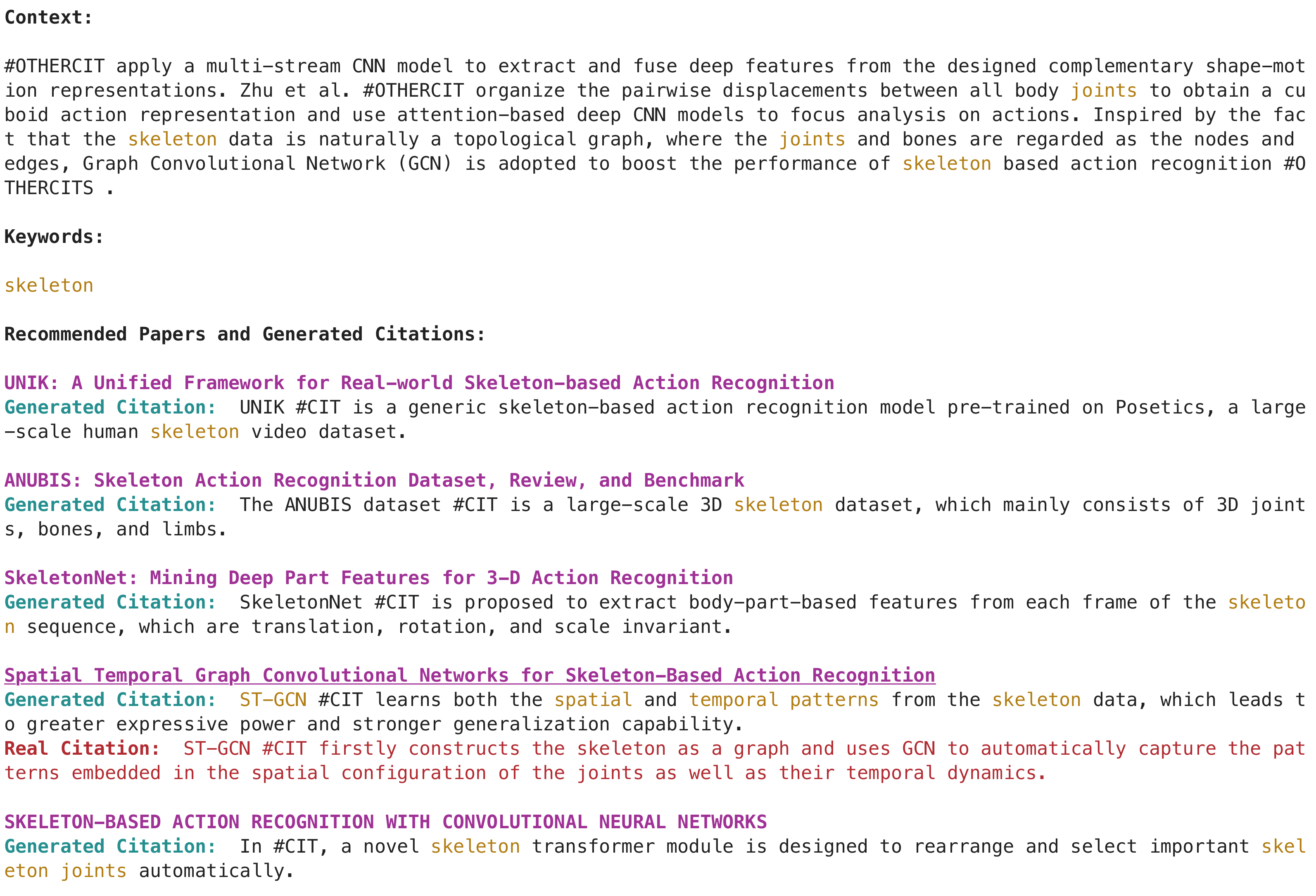}
  \caption{Example of joint paper retrieval and citation generation. The context text was obtained from arXiv. }
  \label{fig:appendix-retrieval-gen-example-1}
\end{figure*}

\begin{figure*}
\centering
  \includegraphics[width=\linewidth]{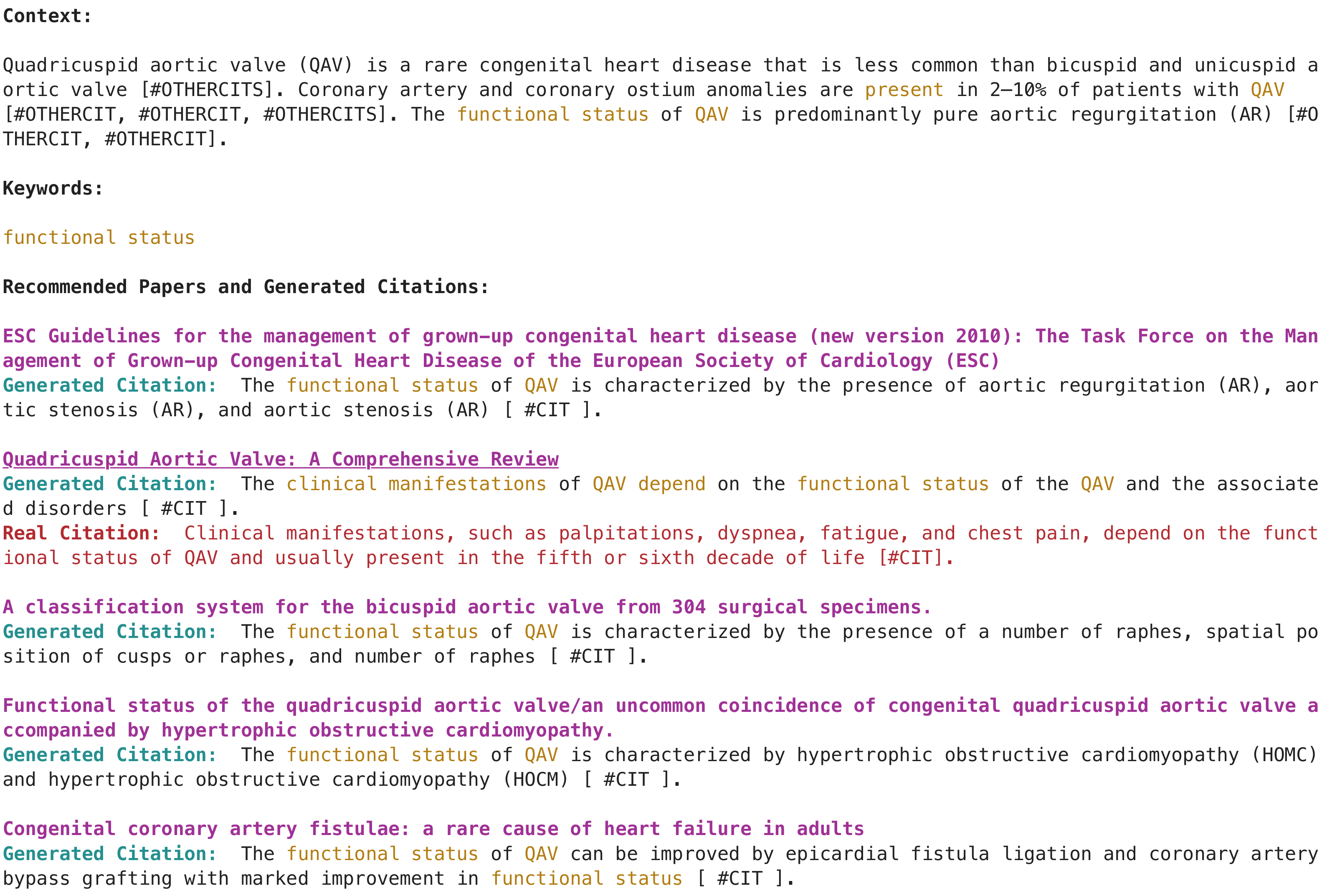}
  \caption{Example of joint paper retrieval and citation generation. The context text was obtained from PMCOA. }
  \label{fig:appendix-retrieval-gen-example-2}
\end{figure*}

We show some specific results of joint paper retrieval and automatic citation sentences generation. We used contexts and keywords that we obtained from papers in arXiv (Figure \ref{fig:appendix-retrieval-gen-example-1}) and PMCOA (Figure \ref{fig:appendix-retrieval-gen-example-2}), respectively. In each example, the actual cited paper occurs in the top 5 paper recommendations, which we have highlighted with an underline.

\end{document}